\documentclass[]{fairmeta}

\title{When to Remember, When to Abstain: Category-Conditioned Retention for Reliable Agent Memory}

\author[1,*]{Olukunle Owolabi}
\author[1, *, \dagger]{Pulkit Gupta}
\author[1, *, \dagger]{Fei Wang}

\affiliation[1]{Meta AI}

\contribution[*]{Work done at Meta}
\contribution[\dagger]{Joint last author}

\abstract{

Persistent agent memory is only as reliable as its retention decision: an
assertion weakly supported by its source can be stored and later reused as
established fact. We study whether the retention decision should be governed by a
confidence bar conditioned on the semantic category of the assertion rather than
by a single global threshold, retaining well-evidenced categories liberally while
abstaining more aggressively where inference is unreliable. We evaluate this in a
deployed cold-start memory pipeline on 100 synthetic personas.
The empirical evaluation is motivated by a sharp reliability asymmetry: across 4{,}715
candidate assertions, only 77.9\% of value and belief assertions are supported by
their source, versus 96.2\% for all other categories. A global confidence
threshold cannot separate these: it either admits unsupported value claims or
discards well-evidenced ones. Conditioning the threshold on category resolves the
tradeoff. In repeated held-out evaluation, a stricter bar on values alone reduces
unsupported retentions from 6.2\% to 4.0\% (an ${\approx}36\%$ relative reduction,
modest but consistent across folds) and, as corroborating evidence, preserves an
estimated 13 percentage points more coverage (95\% CI 9.8--16.0) than a global
threshold at comparable retention.
Our results suggest that reliable retention depends on the type of assertion, not on confidence alone, and that a category-conditioned threshold can act as a simple, effective form of selective prediction at the write boundary.}

\date{\today}
\correspondence{Olukunle Owolabi at \email{oowola01@meta.com}}

\begin{document}

\maketitle

\section{Introduction}
Long-term memory lets agents personalize by accumulating facts about a user across sessions. Most work on agent memory studies organization and retrieval: how memories are structured, updated, and recalled at inference time \citep{packer2023memgpt, park2023generative, zhong2023memorybank, xu2025amem}. Reliability at the \emph{write boundary}, whether a candidate assertion should enter memory at all, has only recently begun to receive attention \citep{zhang2026amac}. Yet this is where reliability is won or lost. A weakly supported assertion that gets written is not just noise; it is later recalled and reused as established fact, and downstream agents act on its stored confidence instead of rechecking the source \citep{kwon2026manufactured}.

In deployed cold-start pipelines, where memory is initialized from a user's existing content before any conversation, the default is to write every extracted assertion (our \textsc{No-filtering} condition). The natural first step toward selectivity is a single global confidence threshold, which we adopt as our comparison point. That threshold assumes confidence means the same thing for every kind of assertion, and we find that it does not. Source support varies sharply by semantic category: assertions about a user's values and beliefs are supported only 77.9\% of the time, against 96.2\% for all other categories. This asymmetry makes a single global threshold ill-posed rather than merely suboptimal.

We treat the write boundary as a selective-prediction problem. A retention policy conditions the confidence bar on semantic category and abstains when an assertion fails the bar for its category. Selective prediction with a reject option is well studied in machine learning generally \citep{hendrickx2024reject}; our contribution is empirical, showing why it matters at the memory-write boundary, through the category-dependent faithfulness asymmetry above, and that acting on it improves faithfulness while retaining more supported content than a global threshold. The policy needs only two signals per assertion, a confidence score and a category, both already produced at extraction time. Because it acts only at the write step and uses signals already produced during extraction, it drops into an existing memory pipeline without changing how memories are stored or retrieved. We instantiate it in a deployed cold-start pipeline and evaluate it on 4{,}715 candidate assertions from 100 synthetic personas.

\section{Related Work}
Prior agent memory systems concentrate on extracting, consolidating, and retrieving information \citep{packer2023memgpt, park2023generative, zhong2023memorybank, xu2025amem}, and increasingly structure it by semantic type, whether by classifying entity types \citep{chhikara2025mem0} or by mining and annotating attributes \citep{salama2025meminsight}.  Memory failures are also well documented and MirageBench \citep{sun2026mirage} shows that personalized LLMs pervasively over-infer user attributes and that confidence is inversely related to fabrication, echoing broader findings that grounded LLMs can be confidently wrong \citep{hagar2025untrue}. 
Both diagnose the problem without acting on it. Recent work on memory admission control gates what is written by scoring candidates on predicted utility, novelty, recency, and a content-type prior \citep{zhang2026amac}. Predicting utility at the write boundary presumes a future query distribution that has not been observed, and it leaves the failure we target untouched, since a fabricated assertion can still be useful. We gate on faithfulness instead, which is decidable from the source already in hand, and we reuse the confidence and category the extractor already emits, so the decision adds no further inference. We show that a single pre-specified category threshold dominates any global threshold at matched retention, and identify the conditions that make this work: unsupported assertions concentrate in one category, and confidence separates supported from unsupported assertions within it.
Selective prediction and abstention are well studied for classification and question answering \citep{hendrickx2024reject, kirichenko2025abstention}; we apply them at the memory-write boundary
\section{Method}
\label{sec:method}

\paragraph{Memory pipeline and write boundary.}
We study a deployed cold-start memory extraction pipeline with a separate extractor per semantic category. Each extractor emits atomic candidate assertions with its own confidence score, so the retention policy reuses an existing model output and needs no additional training or inference. Candidates are deduplicated and then written; no other stage screens them for source support, so the write boundary is otherwise ungated. We insert one decision immediately before persistence: retain the assertion or abstain.

\paragraph{Policies and rationale.}
We compare three policies: writing every candidate, a global threshold applied to all categories, and a category-conditional threshold that may differ by category. The global policy is the special case in which every category receives the same threshold, so the category-conditional family contains it. Category-conditioning helps under two conditions: unsupported assertions must concentrate in one category, and confidence must separate supported from unsupported assertions within it. When both hold, a category-specific threshold removes the weak assertions there without applying the same cutoff to high-support categories. We evaluate the minimal instantiation, with "values" pre-specified as the only gated category and a single fitted threshold, so the held-out comparison tests one stated hypothesis rather than using the same personas to discover a category or fit a full per-category threshold vector.

\paragraph{Data \& Controlled evaluation setup}
We run the pipeline over 100 synthetic personas, producing 4{,}715 candidate assertions. Synthetic personas are deliberate: because each persona's source is complete and known, every assertion can be audited against the evidence the extractor actually saw, which real-user data rarely permits, and no real user content is processed. Thresholds are fit by 20 repetitions of five-fold cross-validation, split by persona: on each fit fold both thresholded policies take the threshold that maximizes faithfulness subject to retaining at least 80\% of candidates, then apply it unchanged to held-out personas. The floor keeps the comparison meaningful, since faithfulness alone is maximized by extreme abstention; what we take to generalize is the ordering of policies, not the particular floor we set. Source support is labeled by our primary LLM judge, Gemini 2.5 Pro (T=0) using fixed rubrics. The judge is stochastic: re-scoring identical inputs flips 4.71\% of verdicts. Appendix~\ref{app:metrics} defines the metrics definitions, and Appendices~\ref{app:faithfulness_rubric}, \ref{app:coverage_rubric}, and ~\ref{app:judge_calibration} give the rubric, coverage, and calibration details

\section{Results}
\label{sec:results}

\begin{figure*}[t]
  \centering
  \includegraphics[width=\textwidth]{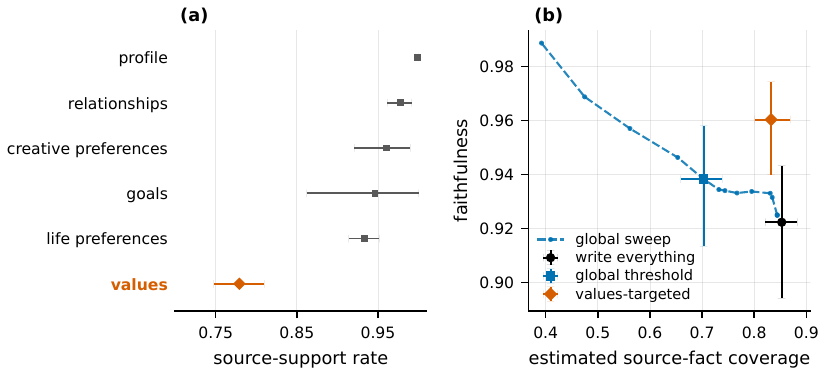}
 \caption{\textbf{The problem and the targeted retention response.}
\textbf{(a)} Source-support rate by semantic category on the evaluation set (Section~\ref{sec:method}); error bars are 95\% persona-clustered bootstrap confidence intervals. Values are the low-support outlier; the single-candidate communication-style category is omitted.
\textbf{(b)} Held-out faithfulness against estimated source-fact coverage. The dashed line is the global-threshold sweep; markers show writing every candidate, the selected global threshold, and the values-targeted threshold, with error bars giving marginal (unpaired) fold intervals. At comparable retention the values-targeted point attains both higher faithfulness and higher coverage than the selected global threshold, and it lies above the global sweep (no global threshold dominates it on both axes); paired policy intervals appear in Table~\ref{tab:heldout_policies}. Post-threshold coverage is estimated, so its magnitude is approximate.}
  \label{fig:headline}
\end{figure*}

\paragraph{Category Separation and Confidence support}
Source support differs sharply by semantic category (Figure~\ref{fig:headline}a). The values category is the low-support outlier at 77.9\% (95\% CI 74.8--81.0), against 96.2\% (95\% CI 95.1--97.1) across all other categories. Values assertions are 21.5\% of candidates but, at a 22.1\% unsupported rate, contribute more unsupported assertions than all other categories combined. Within the values category, support rises from roughly 59\% in the lowest populated confidence bin to roughly 90\% in the highest, so confidence separates supported from unsupported assertions in exactly the category where the failure concentrates.

\paragraph{Targeted abstention outperforms a global threshold.}
Figure~\ref{fig:headline}b shows that no evaluated global threshold matches both the faithfulness and the estimated coverage of the values-targeted policy. On held-out personas the targeted rule reaches 96.03\% faithfulness, 3.79 points above write-all, while estimated coverage falls by 2.01 points. The selected global rule reaches 93.83\% faithfulness, 1.59 points above write-all, but estimated coverage falls by 14.97 points. Relative to the global rule, the targeted policy improves faithfulness by 2.21 points (95\% paired interval 0.22--4.06) and estimated coverage by 12.96 points (9.80--16.04); the coverage difference has the same sign in all 100 folds, though its magnitude is approximate.

The two fitted policies retain comparable fractions of candidates, 80.3\% for values-targeted and 81.8\% for global (paired difference $-1.55$ points, $-5.78$ to $+3.25$), so the faithfulness gain is not explained by more aggressive filtering. Figure~\ref{fig:headline}a identifies the low-support category; Figure~\ref{fig:headline}b and Table~\ref{tab:heldout_policies} show that directing the threshold there beats a single system-wide cutoff.

\begin{table}[t]
\centering
\small
\setlength{\tabcolsep}{3.5pt}
\caption{\textbf{Held-out retention policies.}
Entries are fold means; $\pm$ values are standard deviations. Brackets in the
final row are paired 95\% intervals. Write-all is the fixed full-sample reference. Post-threshold coverage is estimated.}
\label{tab:heldout_policies}
\begin{tabular}{lccc}
\toprule
Policy & Faith. (\%) & Est. cov. (\%) & Retained (\%) \\
\midrule
Write all & 92.24 & 85.27 & 100.0 \\
Global & $93.83\pm1.16$ & $70.30\pm2.09$ & $81.8\pm2.0$ \\
Values-targeted & $96.03\pm0.93$ & $83.26\pm1.73$ & $80.3\pm1.0$ \\
\midrule
Targeted $-$ global
 & $+2.21$ [$+0.22$, $+4.06$]
 & $+12.96$ [$+9.80$, $+16.04$]
 & $-1.55$ [$-5.78$, $+3.25$] \\
\bottomrule
\end{tabular}
\end{table}

\section{Conclusion}
Work on agent memory has largely focused on how memories are organized and retrieved. Our results point to an earlier decision that also affects reliability: which assertions are admitted at the write boundary. When confidence means different things for different kinds of assertion, a single global threshold can be a poor default. If unsupported assertions concentrate in one category while confidence still separates supported from unsupported assertions within it, a type-conditioned threshold can retain more supported content than a uniform cutoff at the same retention. This casts write-time retention as a selective-prediction problem, linking the literature on abstention to a decision that memory systems now make implicitly.
Because the policy needs only a category label and a confidence score that extractors already produce, and changes neither representation nor retrieval, it can be adopted by an existing memory system at low computational cost. More broadly, our findings suggest treating what is not written to memory as a design question in its own right, alongside what to store and how to recall it.

\section{Limitations}
We highlight a few limitations that motivate future work. Faithfulness and coverage are scored by an LLM judge rather than by human annotation; across the judges we tested, the category ordering is stable, but absolute support levels, and the thresholds fit to them, are judge-dependent (Appendix~\ref{app:cross_model} reports cross-model robustness). The controlled evaluation uses a single synthetic corpus, which gives us complete source evidence and lets us measure faithfulness directly; while the pipeline runs in production on real users, the retention approach itself is evaluated only on synthetic data, so validating it on real users, other extractors, and downstream use is the main next step. We also study only the minimal form of the approach: one pre-specified category with a single threshold. Gating at the level of a whole category is coarse, so a miscalibrated threshold or a mislabeled category could systematically drop legitimate content within it; learned per-category thresholds or conditioning on other properties of an assertion are natural extensions that could mitigate this.


\newpage
\bibliographystyle{plainnat}
\bibliography{ref}

\newpage
\appendix
\appendix

\section{Metrics}
\label{app:metrics}
Let $A$ be the set of 4{,}715 extracted candidate assertions, let $\pi_i\in\{0,1\}$ indicate whether policy $\pi$ retains candidate $i$, and let $y_i\in\{0,1\}$ indicate whether the faithfulness judge finds candidate $i$ source-supported. We report retention $Q$ and faithfulness $F$,

$$
Q(\pi)=\frac{\sum_{i\in A}\pi_i}{|A|},
\qquad
F(\pi)=\frac{\sum_{i\in A}\pi_i y_i}{\sum_{i\in A}\pi_i},
$$

where $F$ is the supported fraction among retained assertions. Coverage is recall over a separate population $Z$ of salient source facts:

$$
C_{\mathrm{src}}(\pi)=\frac{1}{|Z|}\sum_{z\in Z}h(z,A_\pi),
$$

where $h(z,A_\pi)$ indicates whether the retained set $A_\pi$ represents source fact $z$. The denominators are distinct: faithfulness and retention are computed over the 4{,}715 candidate assertions (partitioned by semantic category), whereas coverage is computed over the 2{,}634 salient source facts enumerated by the judge (partitioned by salience tier). These are different populations, not subsets, so the counts do not sum.

\section{Faithfulness-Judge Rubric (Abridged)}
\label{app:faithfulness_rubric}

\begin{quote}
\small
Grounding is judged by corroboration, not verbatim entailment. An assertion is grounded if it is stated in the source or is a reasonable inference from converging evidence: several independent signals pointing to the same conclusion, or an explicit profile field. Abstracting a consistent pattern into the interest or value it reveals is legitimate when the evidence converges; asserting a detail with no support is not.

\noindent\textbf{Supported:} the assertion is either (a) directly stated in the source, or (b) a corroborated inference an ordinary reader would draw from converging evidence.

\noindent\textbf{Not supported:} the assertion has no basis in the source, contradicts it, overstates certainty well beyond the evidence, or rests on a single weak or isolated signal. A claim that is merely plausible or thematically consistent, with no specific supporting evidence, is not supported.

\noindent\textbf{Interests vs.\ values:} an interest or preference may be inferred from converging behavioral signals. A principled value or belief is a claim about what the user stands for and needs evidence of the principle itself, such as a stated conviction or a trade-off that demonstrates it. Activities alone do not establish a value; do not upgrade enjoys X into values Y without evidence of the underlying principle.

\noindent\textbf{Internal Calibration:} one explicit statement or profile field is sufficient on its own; an inferred preference, value, or trait requires converging evidence (roughly two or more independent signals); a lone weak signal is not enough.
\end{quote}

For each assertion the judge returns a structured record with the assertion's index and text, a supported or not-supported verdict, and a one-sentence rationale.

\section{Coverage and Salience Rubric (Abridged)}
\label{app:coverage_rubric}
The coverage judge enumerates the salient facts the source supports, tags each with a salience tier, and decides independently whether each is covered by the extracted assertions. It enumerates only source-present facts, marks a fact \textsc{covered} when any candidate expresses it (including by paraphrase, compression, or decomposition) and \textsc{missing} otherwise. HIGH denotes identity-central, durable facts; MEDIUM denotes stable but non-identity-defining preferences or affiliations; and LOW denotes incidental or stylistic detail. Salience is descriptive only: reported coverage is unweighted recall.

The reference set is generated afresh by the judge on every call; it is not a fixed or human-authored gold set, and the dataset's golden-facts field is unused. The stored output records whether a source fact is covered by the full candidate set but not which candidate covers it, so post-threshold coverage relies on a lexical attribution from each covered fact to a candidate and is an estimate. The same estimator is applied to every policy, so coverage differences are directional; their exact magnitude is approximate.

\section{Judge Calibration and Repeatability}
\label{app:judge_calibration}
Calibration cases were labeled by construction: positives were source-derived restatements or paraphrases, and negatives were controlled contradictions, fabrications, or plausible-but-unstated inferences. Boundary cases were resolved by strong-model adjudication and ambiguous ones were excluded; we do not claim a full independently human-labeled calibration set, though an independent human review served as a qualitative alignment bar on the hard cases. On the 67 retained cases (40 easy, 27 hard), taking \textsc{not\_supported} as the positive class, the faithfulness judge reached an F1 of $0.900$ on the hard tier and $0.967$ overall against these constructed labels. We retain the aggregate calibration results but not the per-case adjudications. Separately, re-scoring identical evaluation inputs changed $4.71\%$ of per-assertion verdicts, which we report as an empirical measure of the judge's variability rather than treating it as deterministic.

\section{Cross-Model Robustness of Source-Support Judgments}
\label{app:cross_model}

\begin{table}[h]
\centering
\small
\caption{Source-support rates from three judge configurations on the same 954 assertions from 20 personas (194 values; 760 non-values), with 95\% persona-clustered bootstrap intervals (10,000 resamples).}
\label{tab:cross_model_judges}
\begin{tabular}{lccc}
\toprule
Judge configuration & All & Values & Non-values \\
\midrule
Gemini 2.5 Pro ($T{=}0$) & 0.935 [0.914, 0.954] & 0.840 [0.768, 0.898] & 0.959 [0.939, 0.977] \\
Gemini 2.5 Pro ($T{=}1$) & 0.928 [0.904, 0.951] & 0.845 [0.773, 0.908] & 0.949 [0.927, 0.968] \\
GPT-5 ($T{=}1$) & 0.894 [0.874, 0.915] & 0.624 [0.540, 0.712] & 0.963 [0.952, 0.974] \\
\bottomrule
\end{tabular}
\end{table}

Values is the lowest-support category under every configuration, while non-values stays between 0.949 and 0.963. Temperature has little effect: the paired difference on values ($T{=}1$ minus $T{=}0$) is $+0.005$ [$-0.046$, $+0.056$], with zero near the middle, whereas the cross-model values gap (Gemini $T{=}0$ vs.\ GPT-5) is $+0.216$ [$+0.138$, $+0.295$] on the same assertions. The ordering appears consistent across configurations, and temperature does not seem to drive it, while the absolute magnitude is model-dependent. On non-values the two judges show no detectable difference (cross-model $-0.004$ [$-0.026$, $+0.017$]), so the cross-model difference is concentrated in the values category. All are LLM judges, so this reflects cross-model robustness rather than correctness against human labels.


\end{document}